# Title

- Passively Addressed Robotic Morphing Surface (PARMS) Based on Machine Learning
- Passively Addressed Robotic Morphing Surface.

# Authors

Jue Wang,[1] Michael Sotzing,[1] Mina Lee,[1] Alex Chortos,[1]*

# Affiliations

(1) Department of Mechanical Engineering, Purdue University; 500 Central Dr, Lafayette, IN 47907, USA

*Corresponding author. Email: achortos@purdue.edu.

# Abstract

Reconfigurable morphing surfaces provide new opportunities for advanced human-machine interfaces and bio-inspired robotics. Morphing into arbitrary surfaces on demand requires a device with a sufficiently large number of actuators and an inverse control strategy that can calculate the actuator stimulation necessary to achieve a target surface. The programmability of a morphing surface can be improved by increasing the number of independent actuators, but this increases the complexity of the control system. Thus, developing compact and efficient control interfaces and control algorithms is a crucial knowledge gap for the adoption of morphing surfaces in broad applications. In this work, we describe a passively addressed robotic morphing surface (PARMS) composed of matrix-arranged ionic actuators. To reduce the complexity of the physical control interface, we introduce passive matrix addressing. Matrix addressing allows the control of $N^2$ independent actuators using only 2N control inputs, which is significantly lower than $N^2$ control inputs required for traditional direct addressing. Our control algorithm is based on machine learning using finite element simulations as the training data. This machine learning approach allows both forward and inverse control with high precision in real time. Inverse control demonstrations show that the PARMS can dynamically morph into arbitrary pre-defined surfaces on demand. These innovations in actuator matrix control may enable future implementation of PARMS in wearables, haptics, and augmented reality/virtual reality (AR/VR).

# Teaser

Passive matrix addressing and machine learning control enable under-actuated and highly programmable dynamic shape morphing.

# MAIN TEXT

# INTRODUCTION

Soft actuating systems offer the potential to interface with humans and mimic the behaviors and efficiencies of biological organisms (*1–3*). While traditional rigid actuating assemblies have finite degrees of freedom, soft actuating systems have infinite degrees of freedom, inspiring new strategies to design and control their deformation (*4*). Innovation in soft materials, fabrication techniques, and computational algorithms has led to rapid improvement in the capabilities of soft robots. The emerging rich diversity of shape-morphing strategies has demonstrated value in applications such as implantable medical devices and wearable systems (*5,6*). However, the coordinated movement of large numbers of actuators, which is necessary for applications such as human tactile interfaces and biomimetic robotics, is a nascent field that requires innovations in control strategies and control algorithms.

Robots that morph from one initial shape to one target shape have been developed based on the spatial patterning of materials that deform in response to a global external stimulus (temperature, humidity, etc.), such as liquid crystal elastomers (LCE) (*7, 8*), shape memory alloys (SMA) (*9*), hydrogels (*10, 11*), and shape memory polymers (SMP) (*12, 13*). For these stimulus-responsive materials, the complexity of the shape morphing is determined in the fabrication process. For example, the morphing of a 2D sheet into a 3D shape can be pre-described by depositing stiff 2D structures or multilayers (*14–16*). Similarly, controlling the crease and cut marks of origami (*17, 18*) and kirigami (*19, 20*) structures can define 2D to 3D shape transformations. The design of patterned matter for shape morphing is facilitated by emerging fabrication approaches that include 3D/4D printing (*21–25*), ultraviolet lithography for magnetic particles (*26*), alignment technologies for LCE (*27, 28*), laser or wafer-jet cutting (*20, 29*). For these shape morphing strategies that are prescribed during the fabrication process, the control system can be considered as the modeling that predicts the deformation (*30*). The limitation of this pre-defined shape morphing approach is that only one shape can be formed.

Dynamically programmable robots can morph from their initial shape into many different shapes on demand. This can be accomplished by spatially controlling the stimulus, such as light (*31*), magnetic field (*18*), or temperature (*32*). However, this approach requires an external system to create the spatially varied stimulus, such as an external laser to locally heat the device. Morphing surfaces that do not require external devices have been created using vertical arrays of linear actuators (*3, 18, 33–35*). Independent control of the height of each actuator allows discrete description of the surface. However, these arrays of linear actuators are typically very bulky because a large proportion of the actuator is located underneath the array.

A low-profile actuator array can be created using a surface of bending actuators. Morphing into a target structure can be achieved by adjusting the curvature of each ribbon/beam (*36, 37*). Since the pixels of bending actuators are coupled to each other, the height of each pixel is influenced by the surrounding pixels. This complex coupling leads to a system with infinite degrees of freedom (DoF). Consequently, controlling an infinite DoF with a finite number of control inputs is a key challenge. Control involves two steps: the physical control interface that provides power to each pixel in the array, and the control algorithm that determines how much power to provide to each pixel. To improve low-profile actuator arrays, innovations are required in both the physical control and the algorithms.

Actuator arrays most commonly use direct addressing (*3, 33–35*), which consists of powering each actuator independently using two wires. Controlling an N × N array of actuators with direct addressing requires $N^2$ independent inputs and $2N^2$ wires. As the array size increases, the exponential increase of independent inputs can result in bulky control systems and challenges such as capacitive coupling between the wires. Passive matrix addressing activates individual pixels at overlapping regions of a crossbar array of electrodes (Fig. 1A), which only requires 2N control signals to address a $N^2$ matrix. Passive matrix addressing provides intermittent power to each pixel and has consequently previously been applied to static devices such as sensors and displays (*38–43*). Similarly, a passive matrix robotic system must include pixels that can operate with intermittent power.

While passive matrix addressing provides an intriguing potential physical control interface, control algorithms based on traditional analog control concepts are challenging to implement due to complex mechanical coupling between pixels. The burgeoning machine learning (ML) methods empirically approximate complicated unknown models of systems, and have been widely implemented in sensors (*44–46*), analytical models (*47, 48*) and controllers (*49, 50*) of soft robots. An important step in implementing ML is to obtain a sufficiently large and reliable dataset. Finite Element Modeling (FEM) provides a very reproducible method to generate a large dataset of actuation conditions that is free from non-idealities of real devices, such as hysteresis and manufacturing defects. We have previously introduced the theory behind this combination of FEM and ML to predict the deformation of actuator arrays (*51*).

In this work, we described a passively addressed robotic morphing surface (PARMS) controlled using ML (Fig. 1A and Fig. 1B). Ionic electro-active polymer actuators are used because their capacitive actuation mechanism allows them to retain their actuation state with intermittent power; the ionic actuator can only be actuated when the circuit is a closed loop and maintains its actuation state when the circuit is floating. Given the data from FEM simulations, a trained ML model based on multilayer perceptron (MLP) is implemented to attain model-free control. The ML model can predict the deformation given the voltage applied to each pixel (forward control) with high precision in real-time (at least 50Hz). The ML model also allows the determination of the voltage on each pixel necessary to achieve a target surface (inverse control). Using this ML algorithm for inverse control, PARMS can reproduce arbitrary pre-defined surfaces statically and dynamically, which suggests possible applications in tactile displays and human-machine interactive devices.

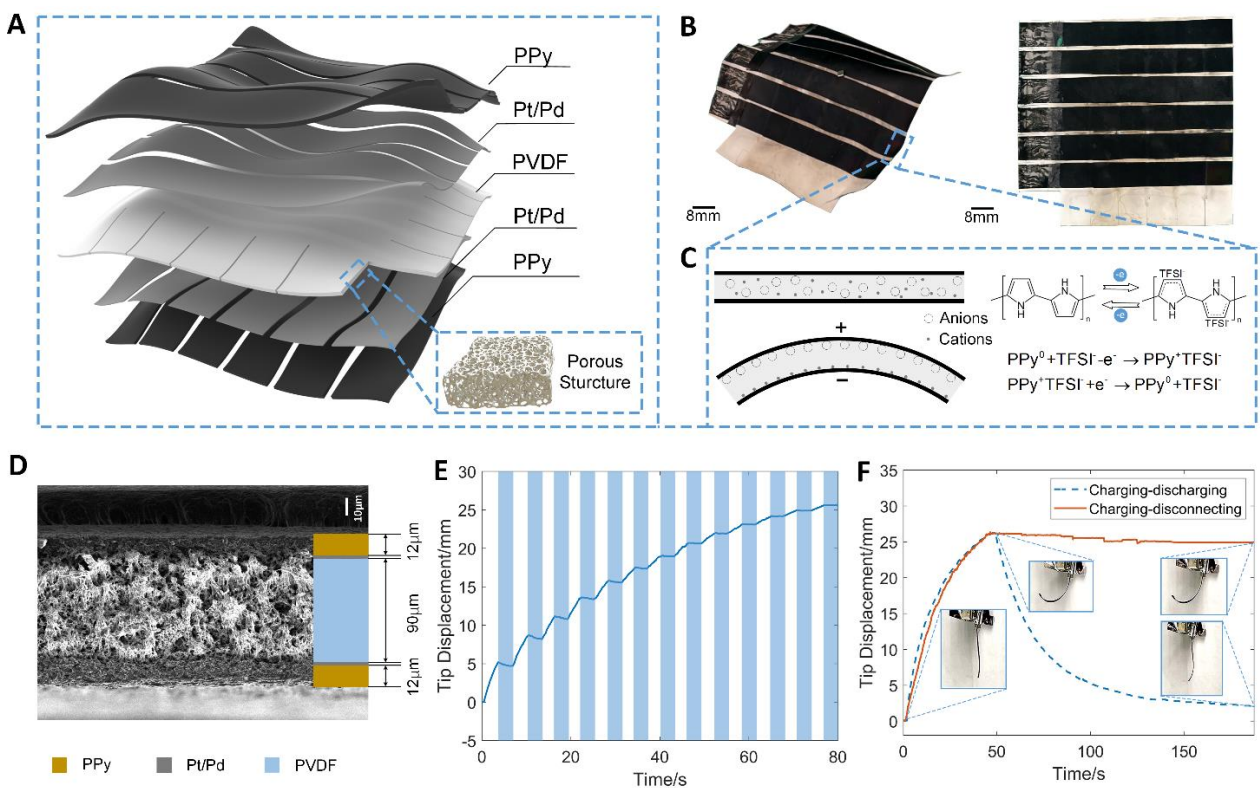

**Fig. 1. The design and principle of the mechanism.** (**A**) Schematic of a passive matrix crossbar array consisting of polypyrrole (PPy) as an active electrode and porous polyvinylidene difluoride (PVDF) as an ionic conductor. (**B**) Image of the PARMS in the actuated state and the original flat state. (**C**) The actuation principle of an ionic electro-active polymer actuator. The PPy at the anode is oxidized and binds the anion $TFSI^-$ while the PPy at the cathode is reduced, losing the anion $TFSI^-$. This causes differential swelling in the PPy electrodes. (**D**) The cross section of ionic actuator under electron microscope. The thickness of substrate and PPy layer are 90 μm and 12 μm, respectively. The thickness of Pt/Pd layers is less than 1 μm. (**E**) The intermittent application of voltage to a ionic actuator strip. The white areas represent charging status (0.8 V) and the light blue areas represent the floating status. (**F**) Comparison between the 'charging-discharging' and 'charging-disconnecting' processes of a ionic actuator strip.

## RESULTS
### Characterization of actuator mechanism and fabrication of array

In a passive matrix, a specific pixel is addressed by activating the corresponding row and column electrodes. Each pixel possesses a unique combination of row and column, but only several pixels can simultaneously be addressed independently (*52*). Consequently, setting the voltage on all pixels in the array requires sequentially addressing the pixels. During the time that a pixel is not being addressed, it is desirable to maintain the actuated state of the pixels. Otherwise, the array will not reach its target quasi-static deformation. Ionic electro-active polymer actuators (hereinafter referred to as ionic actuator) operate as ionic capacitors, which are well suited for energy storage. In this work, the active material of the electrode is chosen as polypyrrole (PPy) because of its well-known fabrication process (*53*). The separator consists of porous poly(vinylidene difluoride) (PVDF) infiltrated with an electrolyte solution composed of lithium bis(trifluoromethanesulfonyl)imide (LiTSFI) in propylene carbonate. In response to an applied voltage, ions migrate to the electrodes. The ion migration causes asymmetric swelling in the two electrodes due to differences in the size of ions and their solvation properties (Fig. 1C) (*54, 55*). The ionic actuator is fabricated by evaporating strips of platinum/palladium (Pt/Pd) thin films onto both sizes of the porous PVDF. The metal thin films are used to electropolymerize PPy onto the electrodes. Fig. 1D shows the cross-section of the device structure. The commercial PVDF membrane is 90 µm while the deposited PPy is about 12 µm. The Pt/Pd layers are difficult to discern since their thickness is less than 1 µm.

The actuation magnitude as a function of applied voltage is quantified for a single strip of ionic actuator using a new definition of tip displacement, the trajectory length of motion (see section S1, Fig. S1-S2, and movie S1). This data informs our finite element model of the actuation (described in the Section "Finite element modeling of PARMS array"). To characterize the ability of an individual ionic actuator to maintain its actuated state when the electrodes are in open circuit (floating state), we apply 0.8 V to a ionic actuator strip intermittently (Fig. 1E). The actuator deforms when the voltage is connected and shows a small amount of spring back in the following floating status. When the actuator is connected to ground after actuation, the actuator discharges in tens of seconds and a residual actuation of only 2.4 mm remains (Fig. 1F). When the actuator is left floating after actuation, the device maintains its actuated state, relaxing by only 4% of the maximum deformation after three minutes (Fig. 1F). To illustrate how this feature enables independent electrical addressing of pixels in an array, we sequentially applied voltage to each pixel in a 1 × 6 array (see section S2, Fig. S4, and movie S2). These pixels actuate when voltage is applied and maintain their actuated state during the actuation of other pixels.

The PARMS array is fabricated by depositing a crossbar array of electrode strips to provide the PARMS with 6 × 6 pixels, which are the primary independent control units. The total size of the actuation area is 54 mm × 54 mm where the actuation pixel is 7.5 mm × 7.5 mm and there is a 1.5 mm gap between each strip (Fig. 1B). While actuating the array, ions migrate across the thickness of the PVDF separator to activate the pixel. Ion migration laterally between pixels contributes to crosstalk that reduces the accuracy of the targeted shape morphing. To limit this horizontal migration of ions, we added polymer blockers in the gaps between electrodes (see Fig. S3 for the fabrication schematic), which fills the pores in the PVDF membrane.

**Passive matrix addressing with progressive scan**

We consider two different protocols for scanning a passive matrix array: direct passive addressing (DPA) and progressive scan (PS). DPA refers to applying a voltage to all rows and columns simultaneously, as shown in Fig. 2A. Progressive scan (PS) refers to setting the voltage on the pixels in the array by scanning the rows of electrodes sequentially. For example, in this step shown in Fig. 2B, the first electrode row is set to 0V while all other rows are left floating. The first row of pixels is actuated, while the rest of the pixels cannot be driven because they cannot draw a current. In the next step, shown in Fig. 2C, only the second-row input is connected to 0V, and all other rows are left floating while the column inputs are set to be the target values of each pixel in the second row. The duty cycle of this PS approach is therefore 1/N, where N is the number of rows. After scanning over all rows, the process will start over again from the first row as a loop (Fig. 2D). In this way, the voltage on each pixel can be set independently but not simultaneously. To determine the refresh time of PS, we tested the current change over time for an actuator (see section S3 and Fig. S5). The current decreases 85.3% in 3 seconds, which we adopt as the time that voltages are applied to each row during PS.

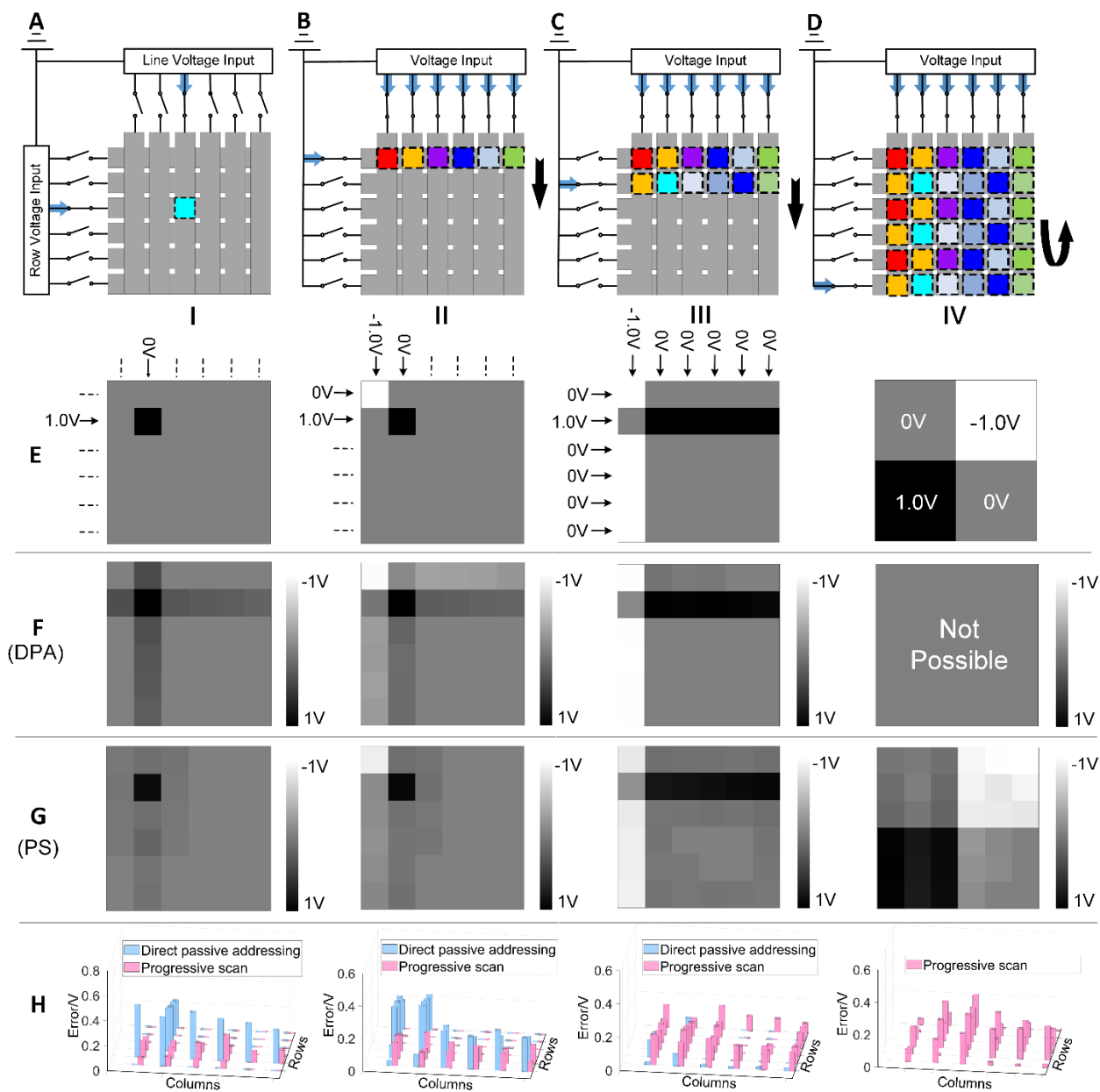

**Fig. 2. Comparison of passive matrix addressing algorithms.** (**A**) The principle of passive matrix addressing using direct passive addressing (DPA) (**B**)-(**D**) The principle of passive matrix addressing using progressive scan (PS). (**E**) 4 demos for illustrating example voltage distributions that may affect the average voltage error. (**E-I**) to (**E-IV**) indicate the ideal voltage distribution. (**E-I**) to (**E-III**) indicate the voltage inputs for DPA. (**E-IV**) indicates the voltage distribution for PS. (**F**) Measured voltage distribution after applying DPA addressing. (**G**) Measured voltage distribution after applying PS addressing. (**H**) The comparison of error distribution. (**H-I**) to (**H-III**) compare the error distributions of DPA and PS using the PARMS with polymer blockers. (**H-IV**) only shows the error distributions of PS since the voltage distribution E-IV cannot be achieved by DPA.

To compare the performance of DPA and PS addressing in our actuator array, we test four demonstration voltage distributions (Fig. 2I-IV). For each of these demos, we measure the voltage error (the error between measured and target voltage on each pixel) of each pixel in the 6 x 6 array. We use the average voltage error over all pixels to quantify the overall error on the voltage in the array, which are shown in Table 1 for all demos. The first demo (Fig. 2E-I) consists of a single pixel at high voltage with all other pixels at zero voltage. The resulting voltage distribution highlights a classic challenge in passive matrix arrays: crosstalk between pixels, which manifests as unintended voltages on adjacent pixels (56, 57). This crosstalk can be caused by parasitic current paths through the electrodes and/or through the ionic separator. Using DPA, the maximum voltage error on an individual pixel is 37.2% (Fig. 2H-I), while the average voltage error over all pixels is 13.4%. By using PS, this average voltage error is reduced to 3.8% (Fig. 2G-I). The second demo (Fig. 2E-II) consists of two pixels with opposite polarity. Using DPA, this is achieved by applying voltage on the first two rows and columns. While the voltage on the two active pixels can be defined exactly, there is significant error on the other pixels of the row and column. Again, the average voltage error is substantially reduced from 15.3% for DPA (Fig. 2F-II) to 4.1% for PS (Fig. 2G-II).

In demo I and demo II, the high average voltage error results from parasitic currents between the floating electrodes. For demo III, a voltage distribution is chosen that does not require any floating rows or columns. Consequently, DPA should be capable of determining the voltage distribution exactly. As expected, the average voltage error exhibits a low value of 2.0%. (Fig. 2F-III) However, for PS, the average voltage error is 9.2% (Fig. 2G-III). This demonstrates that DPA can possess high accuracy when there is no floating input, but this only applies to an extremely small subset of possible voltage distributions. Demo IV is chosen as a voltage distribution that is not possible to produce using DPA. The average voltage error for PS is 11.1% (Fig. 2G-IV).

Based on the observations of demos I-IV, PS can significantly reduce the voltage error when the voltage distribution requires floating connection in DPA. DPA has the lowest voltage error for a very small subset of possible voltage distributions in which all pixels can be programmed simultaneously. Importantly, PS can achieve arbitrary voltage distribution, which is a critical requirement for dynamic shape morphing. Consequently, all demonstrations of shape morphing in this work use PS.

**Table 1. The average voltage error of 4 demonstration voltage distributions under direct passive addressing (DPA) and progressive scan (PS) with and without polymer blockers.**

|  | Demo 1 | | Demo 2 | | Demo 3 | | Demo 4 | |
| --- | --- | --- | --- | --- | --- | --- | --- | --- |
|  | DPA | PS | DPA | PS | DPA | PS | DPA | PS |
| Without blockers | 16.4% | 5.9% | 22.8% | 5.6% | 11.7% | 15.6% | N/A | 33.1% |
| With blockers | 13.4% | 3.8% | 15.3% | 4.1% | 2.0% | 9.2% | N/A | 11.1% |

**Characterizing device non-idealities**

Non-idealities in the array include the ionic coupling between pixels and resistive losses in the electrodes. Adding polymer blockers reduces the parasitic ionic currents, which decreases the average voltage errors (Fig. S6 and Table 1). With the addition of blockers, the average voltage error using progressive scan is reduced from 5.9% to 3.8% for demo I and from 5.6% to 4.1% for demo II. With more active pixels, the impact of polymer blockers becomes more pronounced. In demo III, the average voltage error decreases from 15.6% to 9.2%. In demo IV, adding polymer blockers reduces the average voltage error by

~2/3; from 33.1% to 11.1%. The introduction of polymer blockers reduces parasitic ionic coupling between pixels, realizing approximately 30% − 50% reduction in voltage errors, and its impact increases with the number of active pixels.

Error on the voltage of the pixels can also originate from resistive losses in the electrodes. The voltage applied at the contacts is distributed on the PPy electrodes and across the PVDF dielectric (see Fig. S7A). The low resistance through ionic dielectrics results in meaningful resistive losses in the electrodes. We applied 1V to each pixel of a PARMS array to quantify the voltage drop caused by electrode resistance. The results (see Fig. S7B-S7C) are described in section S2 of the supplementary material. For the pixel furthest from the contacts (with the longest electrodes), 79.6% of the voltage dropped across the device and 20.4% in the electrodes. In the following demonstrations, we have compensated the input voltage based on this result.

**Finite element modeling of PARMS array**

FEM simulations provide a repeatable and low-labor method to collect training data on the actuation of PARMS arrays to train the model for machine learning control. However, when using a computationally-generated dataset to predict the behavior of experimental devices, it is important to ensure that the simulated deformations closely match the experimental devices so that the training data is reflective of experimental conditions.

Abaqus is chosen due to its excellent capability to analyze problems with nonlinear large deformations. However, Abaqus does not include modules relevant to ionic deformation of materials. Consequently, we adopt an approach proposed by Madden et al. (*58*) to approximate ionic deformation using equations for thermal expansion. The strain of the ionic actuator created by ionic transport is proportional to the applied voltage in a particular range of deformation according to the diffusive elastic metal model: $\varepsilon_V = \beta \Delta V$, where $\varepsilon_V$ is the strain indirectly caused by voltage change $\Delta V$ and $\beta$ is the electrical expansion coefficient. The mathematical form of thermal expansion is the same as that of ion-induced strain: $\varepsilon_T = \alpha \Delta T$, where $\varepsilon_T$ is the strain generated by temperature change $\Delta T$, and $\alpha$ is the thermal expansion coefficient. Consequently, the equations for thermal field deformation can be used to approximate the electrical fields. The thermal strain from the resulting simulations indicates the electrically-induced strain in the devices.

As described in the previous section, the electrical resistance of PPy and PVDF can affect the voltage distribution among pixels. These non-idealities are taken into account in our FEM simulations by defining physically realistic values of the electrical/ionic conductivity; in the thermal formulation of the FEM simulations, the electrical conductivity is represented by the thermal conductivity.

To ensure that the simulations closely represent the real devices, the electrical expansion coefficient was calibrated by experimental data on a single actuator (see section S3 and Fig. S8-S10). The $R^2$ value and RMSE of this fitting curve are 0.9977 and 0.0003519, respectively. Additional details of FEM simulation in ABAQUS are shown in section S3. The parameters of the simulation are provided in Table S1. After establishing the simulation parameters, simulations were performed on $6 \times 6$ arrays. The physical array is fixed by a pair of 2 mm × 2 mm magnets at the center so that all four sides are free. Thus, simulations are performed with the same boundary condition (pinned at the center).

**Machine learning based model-free control**

While the combination of progressive scan and polymer blockers enables independent control of the voltage on each pixel, the pixels are intrinsically physically coupled to each other at their boundaries. This physical coupling makes it difficult to establish an explicit analytical model. Hence, we previously proposed a model-free method based on ML to predict the deformation that results from a certain set of input voltages on each pixel (forward prediction) (*51*). The model could also determine the required input voltage on each pixel to achieve a desired surface (inverse design). This method is based on the ML algorithm, the Multi-Layer Perceptron (MLP). The low repeatability of ionic actuators and potential errors in measurement tools could lead to inaccurate datasets that create inaccurate models. Generating the training datasets from aforementioned accurate finite element simulations rather than physical device allows us to create a more accurate model.

To quantify the simulation results as training data for the ML model, we take n × n points uniformly on the surface of the simulation result and vectorize them into vectors of length $n^2$ in order, $\vec{x} = \{x_1, x_2, \ldots, x_{N^2}\}$. Then we vectorize the 6x6 voltage boundary conditions (BCs) in the simulation into a vector of length 36 in the same order, $\vec{y} = \{y_1, y_2, \ldots, y_{36}\}$. As shown in Fig. 3A, when predicting the deformation of the surface (forward control), the voltage BCs are used as the input layer, while the displacement vectors are the output layer. To determine the necessary applied voltages to achieve a target surface (inverse control), the input and output are swapped. We build a fully connected neural network using the MLP model with the hidden layer sizes of (73, 300, 580, 880, 1200) and (901, 700, 550, 300, 180) for forward and inverse control, respectively. The parameters of MLP model are provided in Table S2.

When generating the training datasets, the vectors of applied voltage BCs are defined randomly from -1 to 1. Here, we pre-defined an array, (-1.00, -0.95, -0.90, ..., 0.90, 0.95, 1.00), in MATLAB and randomly picked a value for each pixel. After executing the FEM simulation, the displacement vectors extracted from Abaqus include x, y, and z directions. However, the MLP model only allows the features to be a one-dimensional vector, so we compared the use of z-displacement and total displacement (derived from $\sqrt{x^2+y^2+z^2}$). Theoretically, an exhaustive search of all possible training datasets would require $36^{41}$ simulations. Performing all of these simulations is impractical due to the computational time required. Consequently, we examine the prediction accuracy as a function of the number of training datasets. 100 datasets are used for testing. The R2 score (The closer the R2 score is to 1, the higher training accuracy the model has) and mean squared error (MSE) are shown versus the number of trials in the training data in Fig. 3B and Fig. 3C. The training data must be greater than 1000 to achieve meaningful accuracy. Beyond 1000, the training accuracy continues increasing but begins to saturate. When the number of training simulations is 5000, the R2 score of forward control trained by z-displacement and total displacement are 0.8800 and 0.8728, respectively. For inverse control, these R2 scores are 0.8223 and 0.8550 for the z-displacement and total displacement, respectively. Using total displacement instead of z-displacement improves the accuracy of inverse control but has little effect on forward control. The details of model training are shown in section S4.

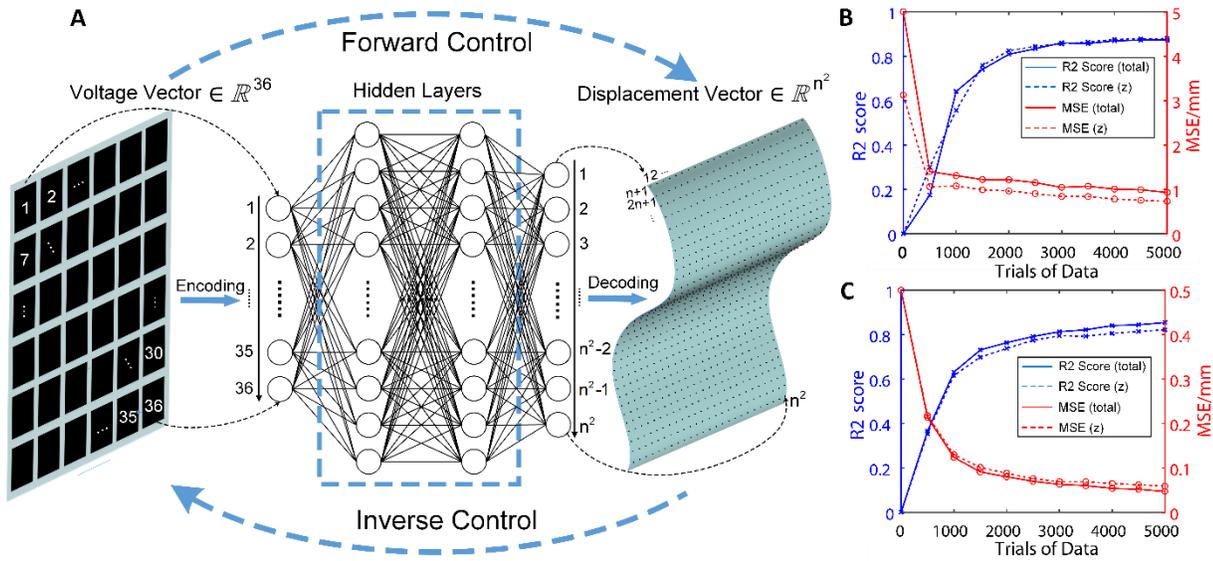

**Fig. 3. The diagram of MLP model and its training properties.** (**A**) The diagram of MLP model. (**B**) The R2 score and MSE versus the size of training data for inverse control. (**C**) The R2 score and MSE versus the size of training data for forward control.

Compared with FEM, which is the most direct way to predict the deformation, the computational burden of executing the ML method is much smaller. The total time for FEM to execute a deformation prediction is about 30 min, while our ML method only requires 0.010 s (see Table S3), which means that online control loops can be executed at a rate of 100 Hz. Real time FEM, up to 60 Hz, has been proposed (*59–61*). However, it is based on simplifying assumptions such as using linear elasticity and a coarse mesh, which decreases the reliability of the prediction. For an actuator array with a large number of actuators, such a course mesh would not be able to accurately predict the deformation.

## Forward control

Forward control predicts the deformation of the surface based on known input voltages. Fig. 4 shows three examples of randomly-generated input voltage arrays. For improved accuracy, the voltage difference between adjacent pixels was constrained to be less than 1V. The deformation video of demo 1 is provided in movie S3.

The simulation results for the three demos are shown in Fig. 4A-F. The labels of the 20 × 20 nodes are first extracted uniformly from the first frame of the simulation data, which is undeformed. In the last frame, the x, y, and z coordinates of these nodes are extracted. Using the least squares method, these points are fitted to a plane and then the points are rotated to ensure the fitted plane is parallel to the x-y plane. At this point, the current z-displacement data of simulation result is ready for further comparison. The ML results are derived from the forward model with the same voltage BCs input. They are vectors of length 400 before decoding. Then we decode them to 20 × 20 matrices and reproduce them in Fig. 4G-I. The data on the deformation of the physical PARMS device is collected from a depth camera and transformed into point cloud (Fig. 4J-L). Then the same fitting and rotation operations are performed on this point cloud. After that, based on the x and y coordinates of the 20 × 20 simulation data, we search in the point cloud to find the closest points to these x and y coordinates (each x-y coordinate of the simulated data corresponds to a point in the point cloud). This way, we can obtain the 400 points in the point cloud with the exact x-y coordinates of the simulation data. Before comparison, we ensure the x-y coordinates of each point in these three datasets are the same and they have the same fitted plane. Therefore, we can convincingly compare the z displacement of these three datasets. Fig. 4M-O depicts the spatial error distribution between real dataset and simulation dataset. The largest errors are close to the edges, and the errors are mainly in the range from -2 mm to 2 mm (Fig. 4P-R). The MSE between real deformation and the FEM simulation results are 0.881 mm, 0.856 mm, and 0.860 mm, respectively. When comparing the data between real mechanism and the ML result, the MSE of these 3 demos are 1.412 mm, 1.217 mm and 1.301 mm, respectively. These errors represent 4.2%, 6.3%, and 6.0% of the maximum height difference in the respective shapes.

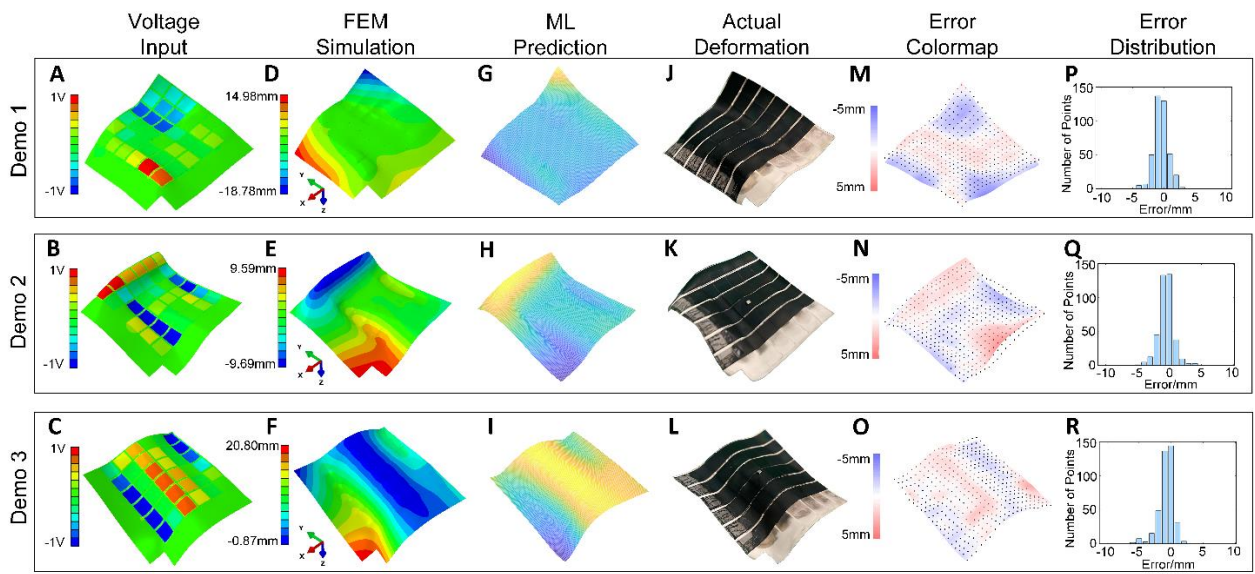

**Fig. 4. Forward control of PARMS using ML.** (**A, B** and **C**) The deformation predicted by FEM with the distribution of the voltage input. (**D, E** and **F**) The deformation predicted by FEM that shows the magnitude of the z-displacement. (**G, H** and **I**) The deformation predicted by the ML algorithm. (**J, K** and **L**) The actual deformation of the PARMS. (**M, N** and **O**) The spatial error distribution between the actual deformation and the simulation result. The black dots are the processed point cloud data collected from physical device and the surface is the derived from simulation result. (**P, Q** and **R**) The numerical distribution of the error.

**Inverse control**

Inverse control generates the input voltage arrays necessary to achieve a target surface deformation. The target surfaces were created in Maya, and were then exported as '*.stl' documents. Meshlab software was used to transfer the '*.stl' documents into point cloud and save them as '*.xyz' documents that include the x-y-z coordinates of each point. Next, Matlab was used to extract the displacement vector from the point cloud as the input to our inverse model. The output voltage arrays were applied to our physical device.

To illustrate dynamic shape morphing capability, we designed two sequences of surfaces as the inputs of the inverse control (Fig. 5). As shown in Fig. 5A, the PARMS started as an undeformed flat sheet, and then the voltage arrays for a sequence of target surfaces were applied. The approximate shape is identifiable, and the shape of PARMS is close to its simulation result. The discrepancies between the target surface and the shape produced through ML and the physical PARMS originates from the limited resolution of the current prototype (6 x 6 pixels). The fourth images from the left depict the spatial error distributions between the simulation result and deformation of physical device. The average error of each step is 0.88 mm, 1.63 mm, 1.84 mm, 1.29 mm, 2.16 mm, and 1.12 mm, respectively. Fig. 5B shows another sequence of shapes with average error: 0.99 mm, 1.57 mm, 1.67 mm, 1.49 mm, 1.59 mm, and 1.04 mm, respectively. (Movie S4, S5)

As a third example of inverse control, the PARMS reproduced the acronym of Purdue University, 'P' and 'U' (Fig. S11 and Movie S6).

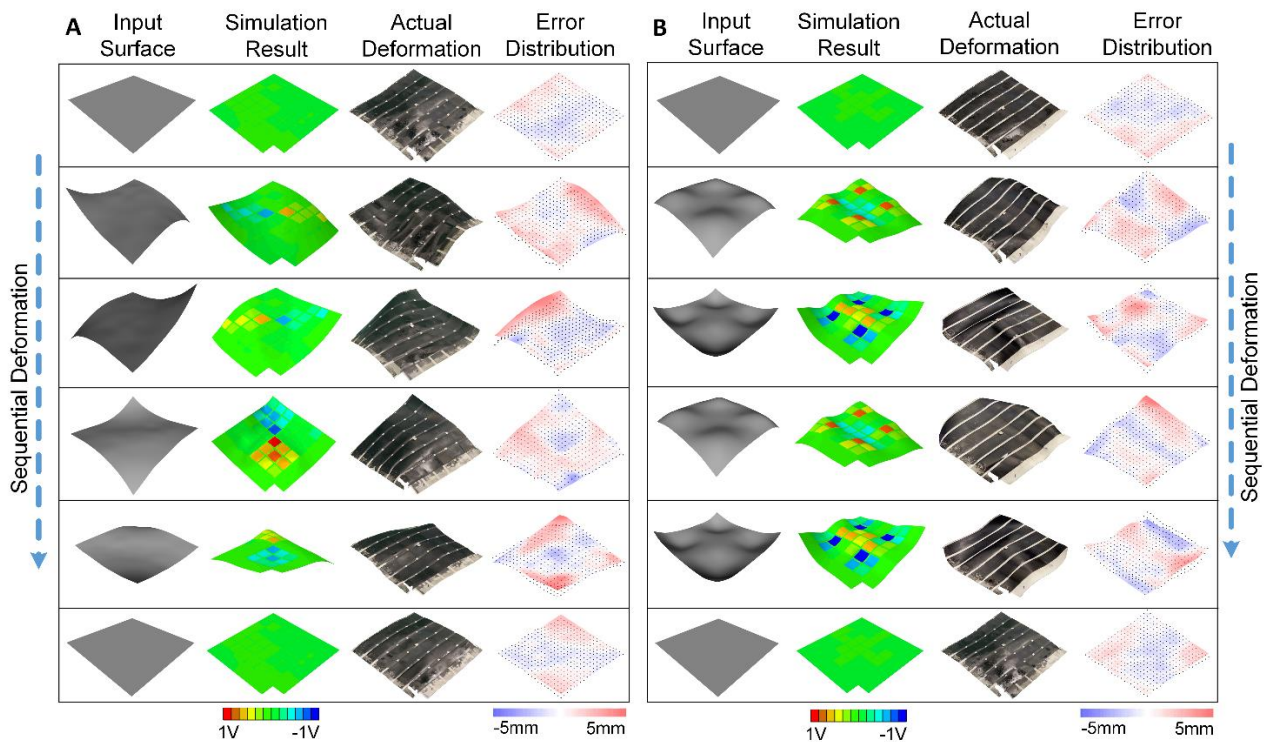

**Fig. 5. Inverse control of dynamic shape morphing series.** (**A**) Demonstration series 1. (**B**) Demonstration series 2. At each step, the first image from left indicates the input target surface, the second image from left indicates the simulated deformation results, the third image from left is an image of the deformation of the physical device, and the fourth image from left shows the error distribution between simulation results and deformation of the physical device. The black dots are the processed point cloud data collected from physical device and the surface is derived from the simulation result.

**Discussion**

As applications of soft actuators extend to arrays of devices with coordinated movements, the physical control interfaces and controller complexity start to become a limitation. Discrete morphing surfaces have the advantage of good programmability, but their linear actuators and direct addressing system ($N^2$ control inputs for $N^2$ controllable pixels) results in bulky devices that are difficult to incorporate into micro-scale systems (*3, 35*). Morphing surfaces composed of bending actuators can be very thin, enabling low-profile device arrays. Recent morphing surfaces that use bending actuators have controlled the curvature of entire rows or columns rather than individual pixels. This limits the independently controllable parameters to 2N, which does not allow the creation of arbitrary shapes (*37, 65*). The low-profile magnetic actuator array described by Bai et al. employed an addressing scheme that is unique to magnetic actuators and allows 4N independent inputs to control the deformation of $N^2$ DoFs (*62*). However, due to the actuation mechanism, coupling between pixels cannot be eliminated. Furthermore, the device requires a large externally applied magnetic field. In our PARMS array, the mechanism to actuate the device operates within the actuator array, without the need for external field sources. In addition, our implementation of passive matrix addressing allows the control of $N^2$ independent pixels using only 2N control signals. Fig. 6 shows the scaling relationship between the control inputs and controllable DoF for different technology platforms. Compared to other approaches, matrix addressing demonstrates a compact control system, which could enable applications in portable and low-profile morphing surfaces.

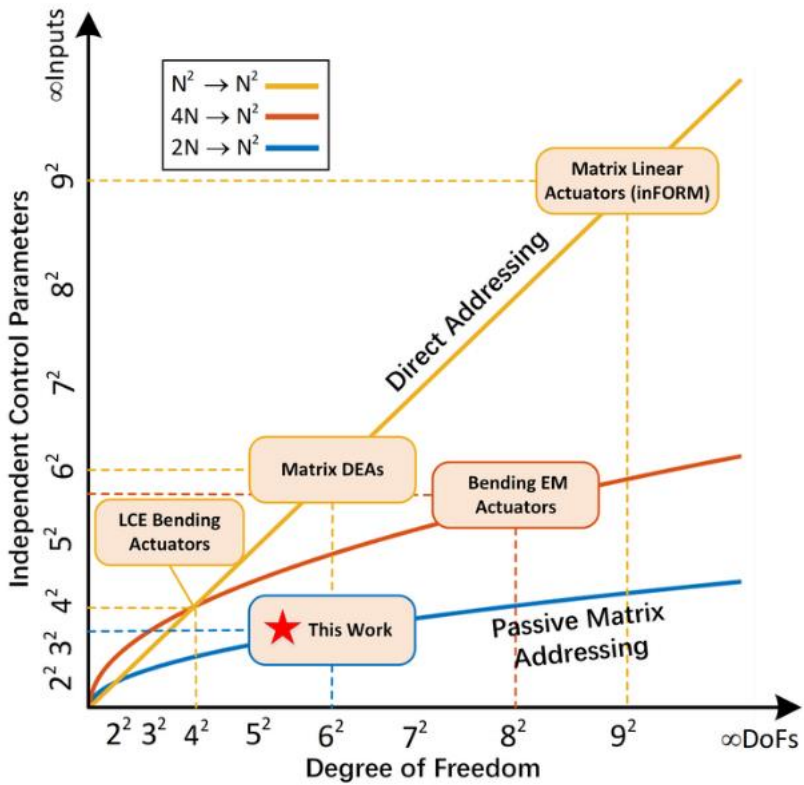

**Fig. 6. Comparison of the independent control inputs vs the degree of freedom (number of actuators) for different actuator array systems.** The Matrix Linear Actuator (InFORM) refers to (*35*). Matrix DEAs refers to (*3*). LCE Bending Actuators refers to (*37*). Bending EM Actuators refers to (*61*).

While this iteration of our PARMS device employs ionic actuators, passive matrix control could enable dynamic shape morphing with any actuator that can store energy (*4*), such as dielectric elastomer actuators and liquid crystal elastomers. In addition to the programmable surfaces described in this work, this matrix addressing approach could be used in applications such as reflective displays (*63*) and microfluidic devices (*64*).

Our use of a control algorithm based on machine learning enables real-time inverse control, which suggests potential suitability for applications such as 3D tangible user interfaces and remote collaboration. For example, virtual surfaces could be valuable for AR/VR devices and remote Human-Computer Interaction (HCI) applications. A virtual surface created in an AR/VR device, such as 3D models, scanned faces, and topographical maps could be displayed on a remote morphing surface in real-time. Two people with the same morphing surfaces could physically interact through the real-time synchronization of two devices.

An important limitation of matrix addressing is the reduced duty cycle at which each pixel is addressed. Using progressive scan, the array can be refreshed at a rate of 1/N of the time it takes to actuate each pixel. Consequently, matrix addressing is most suitable for applications in which it is acceptable for the shape change to occur at speeds lower than the maximum deformation rate of the actuation mechanism.

The ability to reproduce arbitrary surfaces is limited by the $6 \times 6$ array size of the PARMS in this work. Consequently, future research will develop a higher resolution mechanism, which will benefit from innovations in the devices and the ML algorithms. As the array size becomes larger and the pixel size becomes smaller, the coupling between pixels will also become larger. This coupling could be reduced by adding a layer of nonlinear devices (e.g. diodes) into the array (*66, 67*). In addition to improving the array size, it would be advantageous to improve the actuation rate of the array. This could be accomplished by optimizing the thicknesses of the Pt/Pd and PPy electrode layers (*68*) or using electrostatic actuators that operate at higher frequencies.

As the array size increases, larger out of plane displacements could be possible, motivating the development of improved ML controllers. Data collected from simulations includes the x, y, and z displacement of each node. However, MLP only allows one data input for each node (z displacement). Consequently, the training accuracy could be improved by making use of the x and y data. Therefore, some typical models for training point cloud such as 3D convolutional neural network (3D CNN) and PointNet can be potentially applied to train the 3D deformation of the PARMS or other morphing devices.

In summary, we have demonstrated a passively addressed robotic morphing surface (PARMS) that can morph into different 3D shapes on demand. Passive matrix addressing is implemented to control $N^2$ electrically independent outputs using 2N independent inputs, making the system highly under-actuated. Due to mechanical coupling between pixels, we implement machine learning-based model-free control. Forward control demonstrations indicate that the deformation of the PARMS can be predicted with high precision in real-time. Inverse control demonstrations show that the PARMS can dynamically reproduce arbitrary shapes on demand.

# MATERIALS AND METHODS
## Materials and devices

Porous PVDF membranes (Durapore HVLP14250), lithium bis(trifluoromethane), propylene carbonate (99%), pyrrole monomers, and Sylgard 184 KIT were all purchased from Fisher Scientific Co LLC and used as received.

The sputter coater was Cressington 208HR. The microcontroller was Arduino mega2560. The depth camera was Intel RealSense Depth Camera D435. The multi-material DIW printer and electrochemical station were developed in-house.

**Device fabrication**

The details of fabrication are depicted in Fig. S1. First, hydrophilic PVDF membrane with 0.45 μm pores is cut into a 10 cm × 10 cm square. Then, Pt/Pd is sputtered through a shadow mask to deposit six electrode strips on each side of the PVDF membrane to create a crossbar architecture. The sputter coater is operated at a current of 40 mA for 120 s on each side of the PVDF. When depositing the PPy, PPy is deposited on the Pt/Pd electrodes but not the membrane between the electrodes. This generates nonuniform stresses that can cause wrinkles on the membrane. Therefore, we designed a holder to pre-stress and fix the membrane with magnets while it is immersed in the electrochemical workstation. For the electropolymerization of PPy, we prepare a propylene carbonate (PC) solution of 0.1 mol/L pyrrole monomer with 0.1 mol/L $Li^+TFSI^-$ as the electrolyte and add it to a self-made electrolytic bath. After wetting the membrane with the prepared solution, the membrane is pre-stressed and fixed on the holder by four pairs of magnets. The Pt/Pd strips on the membrane are connected to the positive electrode, while a steel wire mesh is connected to the negative electrode. A constant input current of 0.1 mA is used for electropolymerization. To ensure a uniform growth speed on different strips, each strip has an independent input, and it is disconnected from the electrode as soon as the coating process on this strip is finished. Coating is done one side at a time. After finishing the first side, the holder is removed from the electrolytic bath, the membrane is rotated by 90 degrees, and is prestressed and fixed on the holder again. The PPy deposition process is then repeated on the six electrode strips. The total time of electrochemical reaction is approximately 4-5 h. The membrane with PPy on all electrodes is then soaked in a solution of 0.5 mol/L $Li^+TFSI^-$ in PC for another 30 min. Subsequently, the membrane is removed from the holder and washed with water. The membrane is placed between two glass sheets, and external force is applied for 24 h to flatten the array.

Polymer blockers are printed in the gaps between two strips using a direct ink write (DIW) printer. We chose Sylgard 184 with a 10:1 ratio (base to curing agent) as a hydrophobic material that limits ion migration. The composition is dispensed in the gaps between the electrodes using a 0.1 mm nozzle with 20 psi pressure. Finally, the sample is cut to 66 mm × 66 mm and then mounted on the control system.

**Control system**

Fig. S12 illustrates the overall control and measurement system for PARMS. Our mechanism requires a control system with a series of independent channels that generate both negative and positive voltage from -5 V to 5 V. In addition, relays are needed to control the connection to each row and column. Thus, as shown in Fig. S12B, the row inputs are active, generating the specific voltage. We use DAC chips (MCP 4725), converting the digital signal from MCU (Arduino mega2560) to the analog output from 0 V-5 V. The column inputs are passive and are connected directly to a GPIO of MCU that can only output 0 V and 5 V. When the positive voltage is required, it outputs 0 V so that the row inputs minus the column inputs are positive. While the desired voltage is negative, it connects to 5 V so that the difference between the row and column inputs is negative. During the actuation process, positive and negative voltages are set sequentially. The whole process can be described as follows: first, turn on the relay of the first row and relays of line inputs with corresponding positive pixels (if (1,1) and (1,4) require positive voltage, only the relays on line 1 and 4 turn on while the others remain off); after 3s, turn off the first row and turn on the second row; while only turn on the relays of line inputs that its corresponding second-row pixels are positive. The process is then repeated for all rows.  The second round is for pixels with negative voltage, so the row inputs are sequentially connected to the 5V. 0.01 mm copper wire is used to connect the control electronics to the actuator array, which is secured to the mechanism with conductive carbon tape.

**Measurement system**

Measuring the deformation of the actuator array is done using a multi-view stereo-imaging platform consisting of a depth camera (Intel RealSense Depth Camera D435) connected to a computer taking top-view images of the mechanism. The depth camera has two independent webcams inside and calculates depth from stereo vision. It can output point cloud with RGB information with $1280 \times 720$ resolution. Before use, 'Self-Calibration' codes are used in Intel RealSense SDK2.0 to improve the accuracy. After running the on-chip self-calibration routine and storing the result in the flash memory, the depth accuracy improves from $\pm 1.2\%$ to $\pm 0.2\%$. The output of the depth camera is '*.ply' format. The point cloud data is processed to isolate the surface data by transferring the format to '*.stl' using Meshlab and then opening it in Solidworks to manually delete irrelevant data.  Then, the cleaned '*.stl' documents are transferred to '*.xyz' format, which is converted to '*.txt' format that contains the 3D position and orientation information of each point cloud.

**Finite Element Simulations and Machine Learning Model**

The finite element modeling process is described in Section S3 and Table S1 in the supplementary information. The data preparation and machine learning algorithm are described in Section S4, Table S2, and Table S3 in the supplementary information.

**Acknowledgments**

This research was supported by the startup funding at Purdue University. We thank the school of agriculture of Purdue University for providing sputter coater and electron microscope for the fabrication of the PARMS and thank the school of mechanical engineering of Purdue University for providing electrical measurement devices. We thank Ruhaan Joshi for designing a multi- material direct ink writing printer that is used for printing polymer blockers of PARMS. We thank Pranav Parigi for testing the electrical property of ionic actuator. Author contributions: Jue Wang and Alex Chortos proposed the idea of this paper. Jue Wang designed the PARMS with its control system, conducted the experiments, collected the data and edit the manuscript. Alex Chortos directed the design and experimental process. Michael Sotzing contributed to the fabrication of PARMS and reviewed the manuscript. Mina Lee contributed to the material optimization and reviewed the manuscript.

**Funding:**
   Startup funding at Purdue University.

**Author contributions:**
   Conceptualization: JW, AC
   Methodology: JW, AC, MS
   Investigation: JW, AC, ML
   Visualization: JW
   Funding acquisition: AC

Project administration: AC
Supervision: AC
Writing – original draft: JW
Writing – review & editing: AC, MS, ML

**Competing interests:** The authors declare that they have no competing interests.
**Data and materials availability:** All data needed to evaluate the conclusions in the paper are presented in the main manuscript and/or the supplementary materials. Additional data are available from the authors upon request.

**Figures and Tables**

**Fig. 1. The design and principle of the mechanism.** (**A**) Schematic of a passive matrix crossbar array consisting of polypyrrole (PPy) as an active electrode and porous polyvinylidene difluoride (PVDF) as an ionic conductor. (**B**) Image of the PARMS in the actuated state and the original flat state. (**C**) The actuation principle of an ionic electro-active polymer actuator. The PPy at the anode is oxidized and binds the anion $TFSI^-$ while the PPy at the cathode is reduced, losing the anion $TFSI^-$. This causes differential swelling in the PPy electrodes. (**D**) The cross section of ionic actuator under electron microscope. The thickness of substrate and PPy layer are 90 μm and 12 μm, respectively. The thickness of Pt/Pd layers is less than 1 μm. (**E**) The intermittent application of voltage to a ionic actuator strip. The white areas represent charging status (0.8 V) and the light blue areas represent the floating status. (**F**) Comparison between the 'charging-discharging' and 'charging-disconnecting' processes of an ionic actuator strip.

**Fig. 2. Comparison of passive matrix addressing algorithms.** (**A**) The principle of passive matrix addressing using direct passive addressing (DPA) (**B**)-(**D**) The principle of passive matrix addressing using progressive scan (PS). (**E**) 4 demos for illustrating example voltage distributions that may affect the average voltage error. (**E-I**) to (**E-IV**) are the ideal voltage distribution. (**E-I**) to (**E-III**) indicate the voltage inputs for DPA. (**E-IV**) indicates the voltage distribution for PS. (**F**) The voltage distribution for DPA on the PARMS. (**G**) The voltage distribution for PS on the PARMS. (**H**) The comparison of error distribution. (**H-I**) to (**H-III**) compare the error distributions of DPA and PS using the PARMS with polymer blockers. (**H-IV**) only shows the error distributions of PS since the voltage distribution E-IV cannot be achieved by DPA.

**Fig. 3. The diagram of MLP model and its training properties.** (**A**) The diagram of MLP model. (**B**) The R2 score and MSE versus the size of training data for inverse control. (**C**) The R2 score and MSE versus the size of training data for forward control.

**Fig. 4. Forward control of PARMS using ML.** (**A, C** and **E**) The real deformation of the PARMS. (**B, D** and **F**) The deformation predicted by the ML algorithm. (**G** to **L**) The deformation predicted by FEM, where **G**, **I**, and **K** show the distribution of the input voltage and **H**, **J**, and **L** show the magnitude of the z-displacement. (**M** to **R**) The error distribution between the real deformation and the simulation result where

**M**, **O** and **Q** are the numerical distribution and **N**, **P** and **R** are the spatial distribution.

**Fig. 5. Inverse control of dynamic shape morphing series.** (**A**) Demonstration series 1. (**B**) Demonstration series 2. At each step, the first image from left indicates the input target surface, the second image from left indicates the simulated deformation results, the third image from left is an image of the deformation of the physical device, and the fourth image from left shows the error distribution between simulation results and deformation of the physical device. The black dots are the processed point cloud data collected from physical device and the surface is derived from the simulation result.

**Fig. 6. Comparison of the independent control inputs vs the degree of freedom (number of actuators) for different actuator array systems.** The Matrix Linear Actuator (InFORM) refers to (*35*). Matrix DEAs refers to (*3*). LCE Bending Actuators refers to (*37*). Bending EM Actuators refers to (*61*).

**Table 1. The average voltage error of 4 demonstration voltage distributions under direct passive addressing (DPA) and progressive scan (PS) with and without polymer blockers.**

**Supplementary Materials**

Section S1. Preliminary experiment of uniform strip ionic actuator
Section S2. Characterization of the matrix addressing
Section S3. Details of simulation
Section S4. Details of training MLP model
Fig. S1. The diagram of the definition of tip displacement
Fig. S2. The deformation of a strip ionic actuator
Fig. S3. The fabrication process of the PARMS
Fig. S4. The decoupling characteristics of a 1x6 mechanism under passive matrix addressing
Fig. S5. Current change over time for an ionic actuator
Fig. S6. The effect of polymer blockers on the voltage distribution on the PARMS
Fig. S7. The actual voltage drop of each pixels when the PARMS is driven passively
Fig. S8. The comparison between experiment result and simulation result of the strip ionic actuator
Fig. S9. The calibration of the thermal expansion coefficient
Fig. S10. Four samples of the simulation of PARMS that are used for MLP training
Fig. S11 Inverse control of quasi-static deformation of PARMS.
Fig. S12. The control and measurement system of the mechanism
Table S1. The parameters of the finite element simulation
Table S2. The parameters of MLP model
Table S3. The time of training and executing models
Movie S1. Actuation properties of single strip ionic actuator
Movie S2. The demonstration of the PARMS's decoupling feature
Movie S3. Forward control of PARMS compared with simulation and ML prediction

Movie S4. Dynamic deformation by inverse control of PARMS (Demo 1)
Movie S5. Dynamic deformation by inverse control of PARMS (Demo 2)
Movie S6. The deformation of letters by inverse control of PARMS